\documentclass[11pt]{article}

\PassOptionsToPackage{table}{xcolor}
\usepackage[preprint]{acl}
\usepackage{times}
\usepackage{latexsym}
\usepackage[T1]{fontenc}
\usepackage[utf8]{inputenc}
\usepackage{microtype}
\usepackage{inconsolata}

\usepackage{graphicx}
\usepackage{booktabs}
\usepackage{amsfonts}
\usepackage{amsmath}
\usepackage{algorithm}
\usepackage{algorithmic}
\usepackage{fvextra}

\newcommand{\nf}{NF}
\newcommand{\mpdrel}{RelScore}
\newcommand{\mpdthree}{RoleScore}
\newcommand{\scanswer}{RepeatAnswer}
\newcommand{\scscoring}{RepeatScore}
\newcommand{\screl}{RepeatRel}
\newcommand{\qualityonce}{QualityOnce}
\newcommand{\relonce}{RelOnce}

\title{A Single Pass Captures the Gain of Document Assessment\\ in Retrieval-Augmented Generation}

\author{Jungseob Lee\textsuperscript{1} \quad Sugyeong Eo\textsuperscript{2} \quad Chanjun Park\textsuperscript{3} \\
\bfseries Hyeonseok Moon\textsuperscript{4}{\hypersetup{linkcolor=black}\thanks{Corresponding authors.}} \quad Jaehyung Seo\textsuperscript{5}\footnotemark[2] \\[4pt]
{\normalsize\mdseries \textsuperscript{1}Korea University \quad \textsuperscript{2}Yonsei University Mirae Campus \quad \textsuperscript{3}Soongsil University} \\
{\normalsize\mdseries \textsuperscript{4}Sookmyung Women's University \quad \textsuperscript{5}Konkuk University} \\[2pt]
{\small\mdseries \texttt{omanma1928@korea.ac.kr} \quad \texttt{s.eo@yonsei.ac.kr} \quad \texttt{chanjun.park@ssu.ac.kr}} \\
{\small\mdseries \texttt{hyns.moon@sookmyung.ac.kr} \quad \texttt{seojae777@konkuk.ac.kr}}}


\definecolor{ourblue}{rgb}{0.368,0.507,0.71}
\definecolor{methodblue}{HTML}{EAF3F9}
\definecolor{groupyellow}{HTML}{FFF6DC}
\hypersetup{
    colorlinks,
    linkcolor=ourblue,
    citecolor=ourblue,
    urlcolor=ourblue
}

\begin{document}
\setcounter{footnote}{1}
\maketitle
\begin{abstract}
Retrieval-augmented generation gives a large language model (LLM) only a few retrieved documents, so outdated or misleading passages can displace useful evidence. Document assessment scores the candidates before generation, and recent systems spend extra LLM calls on repeated judgments or specialized assessor roles. However, it is unclear how much of this work a system with a fixed retriever and answer generator needs. To answer this question, we compare assessment recipes under a matched protocol that fixes the questions, candidate pools, answer generator, and context budget. Across conflicting, entity-ambiguous, and open-domain question answering with five open-weight LLMs, we find that a single assessment pass, such as one relevance judgment for each document, captures the gain of assessment, whereas repeated judgments, extra roles, and answer voting multiply the calls without a reliable further gain. The gain comes chiefly from which documents enter the context and tends to be largest under tight budgets. For new retrieval settings without answer labels, we propose a score-entropy pilot that recommends which recipe to validate first, and on two held-out benchmarks it recommends the most accurate recipe or one nearly tied with it.
\end{abstract}

\section{Introduction}

Retrieval-augmented generation (RAG) supplies a large language model (LLM) with retrieved passages that can support its answer~\citep{lewis2020retrieval}. The answer context holds only a few documents. However, a retrieved list can also contain outdated statements, unsupported claims, or passages about a different entity~\citep{chen2024benchmarking,cuconasu2024power,cattan2025dragged}, and each such passage occupies a slot that useful evidence could have filled.

\begin{figure*}[t]
\centering
\includegraphics[width=\textwidth]{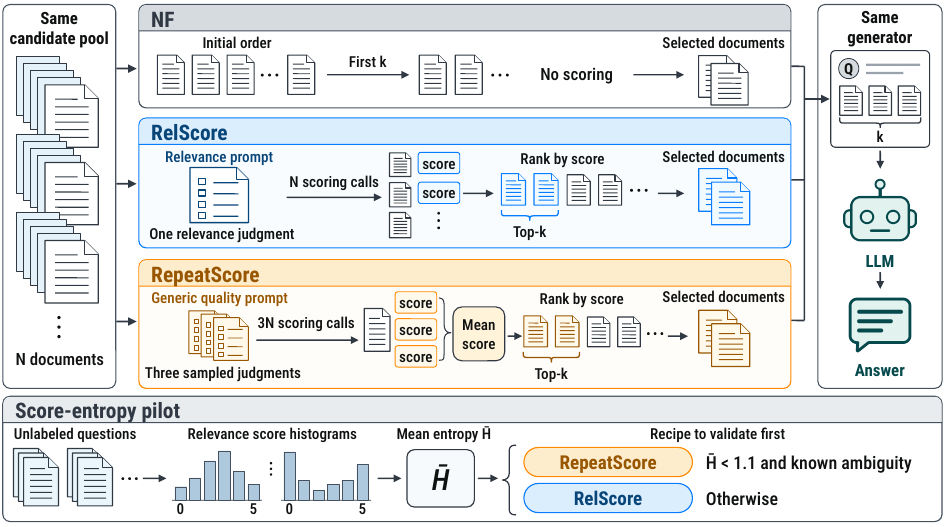}
\caption{Matched comparison of document-assessment recipes. All recipes share the candidate pool, answer generator, and context budget, and an unlabeled score-entropy pilot recommends which recipe to validate first.}
\label{fig:overview}
\end{figure*}

Document assessment addresses this problem by judging or reranking the candidates before generation~\citep{asai2024selfrag,yu2024rankrag,yan2024crag,chang2025mainrag,sun2023rankgpt}. Recent systems go further and spend more LLM calls on this step. For example, they assign specialized agents to evaluate the documents or let agents debate conflicting evidence~\citep{wang2025madamrag,hu2025drag,smit2024going,lee2026madara}, and sampling several outputs offers another way to spend calls~\citep{wang2023selfconsistency,lee2026directional}. Each addition multiplies the number of LLM calls. Moreover, the value of the extra calls is difficult to isolate because published systems differ in their retrievers, answer prompts, and context sizes. We therefore ask how much document assessment a system with a fixed retriever and answer generator needs.

In this work, we answer this question with a matched protocol, illustrated in Figure~\ref{fig:overview}. Specifically, every assessment recipe receives the same questions, the same ordered candidate pools, the same answer generator and prompt, and the same number of documents in the answer context. The unfiltered baseline, \nf{}, sends the first documents of the initial order, whereas \mpdrel{} asks an LLM assessor for one relevance score for each document and sends the highest-scoring ones. We compare this single assessment pass with recipes that spend more LLM calls. \scscoring{} averages three sampled judgments from a generic quality instruction, and further variants repeat the relevance judgment, add consistency and conflict assessors, or vote over sampled answers. To separate the effect of the instruction from that of the repetition, we also evaluate a single sampled judgment of each instruction.

Across three question-answering benchmarks and five open-weight LLMs, we find that a single assessment pass captures the benefit of assessment. Pooled over the models, one relevance judgment for each document reliably improves on unfiltered retrieval on every benchmark. In contrast, repeating judgments or adding assessor roles takes two to three times as many assessment calls and brings no reliable further gain, and voting over answers without reassessing the documents does not help. On open-domain questions, three quality judgments are slightly more accurate than one relevance judgment, but this lead does not come from the repetition, because averaging three sampled judgments never reliably improves on a single judgment of the same instruction. Furthermore, we show that the benefit of assessment comes chiefly from which documents enter the context. This benefit tends to be largest when few documents fit in the context, and it remains when the selected documents are sent in their initial order rather than in score order.

These findings make a single assessment pass a strong default, but a matched comparison in a new retrieval setting still needs answer labels. Therefore, we also examine an unlabeled score-entropy pilot that uses the relevance scores of a small sample of candidate pools to recommend which recipe to validate first. When most candidates share a few score levels, the relevance scores separate them only coarsely, and the pilot recommends a recipe that scores them with another instruction. A labeled comparison then makes the final choice. On the three primary benchmarks, the relevance scores spread over enough levels for the pilot to keep the single relevance judgment. On two held-out benchmarks, the pilot instead recommends three quality judgments, and we find that this recipe is the most accurate or nearly tied with the most accurate one. Even on these benchmarks, repeating the quality judgment adds no reliable gain over a single one.

Our contributions are as follows.
\begin{itemize}\setlength{\itemsep}{1pt}\setlength{\parskip}{0pt}
\item We design a matched protocol that isolates document assessment from the retriever, the answer generator, and the context budget.
\item We show that a single assessment pass, such as one relevance judgment for each document, captures the benefit of assessment across benchmarks and model families, while repeating judgments, adding assessor roles, or voting over answers adds calls without a reliable further gain.
\item We trace this benefit to document selection under a limited context, using budget sweeps and a control that keeps the selected documents in their initial order.
\item We propose a score-entropy pilot that recommends which assessment recipe to validate first for a new retrieval setting without answer labels.
\end{itemize}

\section{Related Work}

\paragraph{Assessing retrieved evidence.}
Existing methods assess retrieved evidence by changing the generator or the pipeline around it. Self-RAG learns reflection signals for retrieval and generation~\citep{asai2024selfrag}, while RankRAG unifies context ranking and answer generation through instruction tuning~\citep{yu2024rankrag}. Corrective RAG uses an evaluator to identify retrieval problems~\citep{yan2024crag}, and MAIN-RAG filters documents using model token probabilities~\citep{chang2025mainrag}. Robust Fine-Tuning trains a generator to detect retrieval defects and extract useful evidence~\citep{tu2025robust}, and OpenDecoder incorporates document-quality indicators into learned attention computations~\citep{mo2026opendecoder}. Unlike these methods, we hold the retriever and the answer generator fixed and compare only how an existing candidate pool is assessed.

\paragraph{Ranking and additional inference.}
Cross-encoders~\citep{nogueira2019passage} and listwise LLM rankers~\citep{sun2023rankgpt} score or order passages before generation. Multi-agent methods assign specialized roles or compare interpretations of conflicting evidence~\citep{wang2025madamrag,hu2025drag,nguyen2025marag}, and self-consistency aggregates sampled answers~\citep{wang2023selfconsistency}. Closest to our setting, \citet{lee2026madara} find that for weaker answer generators, answering from each document in isolation can match full multi-agent assessment, and they use a label-free perturbation probe to route each model to a treatment. Our study instead keeps one joint answer context for every recipe and varies only which documents enter it and in what order. While their probe characterizes a model, our pilot characterizes a candidate-pool construction from the scores of one fixed assessor.

\paragraph{Computation routing.}
Prior routing methods use retrieved-context statistics or short model drafts to decide how much computation a query receives~\citep{wang2025skewroute,wang2025targ,lee2026dart_routing}. In contrast to these query-level methods, our pilot recommends one assessment recipe for a whole retrieval setting. It computes its entropy from integer relevance scores and therefore needs neither token probabilities nor a preliminary answer.

\section{Document-Assessment Protocol}
\label{sec:method}
We compare all recipes through one selection interface, so that assessed recipes differ only in the scores they assign to the same candidates. We define this interface first and the recipes next.
\subsection{A Shared Selection Interface}
For a question $q$, let $\mathcal D_q=(d_1,\ldots,d_N)$ be its candidate pool in the initial order and $k$ the number of documents that fit in the answer context. No filtering, denoted \nf{}, sends the first $\min(k,N)$ candidates. An assessed recipe assigns a score $S_i$ to each candidate and selects the $\min(k,N)$ highest-scoring ones, breaking ties by the initial order. The selected documents are presented in descending score order, and assessor scores or explanations never enter the answer prompt.

Each comparison fixes the questions, the candidate texts and their initial order, the answer generator and prompt, the truncation of each passage, and $k$. Consequently, assessment can change the answer only through which documents are selected and the order in which they appear. When all candidates fit, only the order can change. To separate the two, we add a selection-only control that sends the documents chosen by an assessed recipe in their initial order. Appendix~\ref{app:algorithm} gives the complete procedure.

\subsection{Assessment Recipes}
\paragraph{One relevance judgment.}
\mpdrel{} asks an LLM assessor to rate how relevant each document is to the question on an integer scale from 0 to 5, from irrelevant text to a passage that directly answers the question. The returned relevance score $r_i$ is the selection score, $S_i=r_i$, and the judgment is decoded greedily. The assessor also returns a short quote or paraphrase as evidence, which only the role recipe below uses.

\paragraph{Repeated judgments.}
\scscoring{} samples a generic document-quality instruction three times at temperature 0.7 and selects documents by their mean score,
\begin{equation}
 S_i=\frac{1}{3}\sum_{j=1}^{3}s_i^{(j)}.
\end{equation}
\scscoring{} differs from \mpdrel{} in its instruction, its sampled decoding, and its averaging of three judgments, so we add three controls that change one of these factors at a time. \screl{} samples the relevance instruction three times at the same temperature, while \relonce{} and \qualityonce{} each keep a single sampled judgment of relevance and of quality, respectively.

\paragraph{Additional assessor roles.}
\mpdthree{} tests whether judging documents from further perspectives improves on relevance alone. A consistency assessor scores each document against the evidence extracted from the other documents. One joint conflict call reads the evidence of all documents and assigns each an adjustment that favors majority-supported and recent information. The three judgments are combined as
\begin{equation}
 S_i=0.5r_i+0.3u_i+0.2c_i,
\end{equation}
where $u_i$ is the consistency score and the conflict score $c_i$ is the mean of $r_i$ and $u_i$ shifted by the adjustment of document $i$ and clipped to the same 0--5 scale. The weights favor relevance, and other combinations of the same scores select nearly the same documents, as Appendix~\ref{app:k_sensitivity} shows.

\paragraph{Answer voting.}
\scanswer{} spends its extra calls on the answer instead of the documents. It samples three answers from the unchanged \nf{} context and returns the most frequent normalized answer, following self-consistency~\citep{wang2023selfconsistency}.

\paragraph{Inference work.}
We measure inference work in model calls, because batching and parallel assessment make latency depend on the hardware. With $N$ candidates, \mpdrel{}, \relonce{}, and \qualityonce{} make $N$ assessment calls, \mpdthree{} makes $2N+1$, and \screl{} and \scscoring{} make $3N$. Every recipe then makes one answer call, except \scanswer{}, which makes three answer calls and no assessment call.

\section{Experimental Setup}
\label{sec:setup}

\paragraph{Candidate pools.}
We use three primary benchmarks whose candidate pools pose different problems, namely conflicting web evidence, entities that share a name, and ordinary retrieval noise. All pools are fixed before assessment, and every recipe in a comparison receives the same pools in the same initial order. CONFLICTS~\citep{cattan2025dragged} pairs each question with Google Search results collected by its authors, each reduced to its most question-relevant window of 512 tokens and dated when a date is available. We keep the released order and evaluate the 237 of its 458 questions that have a single annotated answer. RAMDocs~\citep{wang2025madamrag} supplies 500 pools in which evidence for entities that share a name is mixed with misinformation and unrelated noise. Its released order places a correct document first in 476 pools. For NQ Open~\citep{kwiatkowski2019natural}, we query English Wikipedia search with the question, split the returned pages into passages, and keep up to ten passages ranked highest by a BM25-style lexical score~\citep{robertson2009bm25}, which also defines their initial order. No assessor receives gold answers or evidence labels.

\paragraph{Controlled distractors.}
To vary only the kind of distractor, we build controlled pools for 500 TriviaQA questions~\citep{joshi2017triviaqa}. Each pool combines up to five supplied evidence passages with five injected passages. One condition injects passages generated to support a plausible wrong answer, and the other injects evidence passages of other questions. Positions are shuffled once, and every recipe sees the same pool. Scoring and answering read the first 1{,}000 characters of each passage.

\paragraph{Held-out benchmarks.}
We evaluate the unlabeled pilot on two further benchmarks that play no role in the primary comparisons. MuSiQue~\citep{trivedi2022musique} supplies multi-hop questions with paragraph pools, which we keep in their released order. For AmbigQA~\citep{min2020ambigqa}, we build ten-document pools from the annotated Wikipedia pages and Wikipedia search results and shuffle them once. Each benchmark contributes 500 questions, and the relevance scores of the first 50 form its pilot.

\paragraph{Models and decoding.}
Mistral-7B-Instruct-v0.3~\citep{jiang2023mistral} appears in every comparison and runs all controls. The main comparison adds Llama-3.1-70B-Instruct~\citep{dubey2024llama3} with AWQ quantization~\citep{lin2023awq}, Qwen3-4B and Qwen3-8B~\citep{yang2025qwen3}, and DeepSeek-R1-Distill-Llama-8B~\citep{guo2025deepseek}, and the same checkpoint assesses and answers within each comparison. We choose these models to cover 4B to 70B parameters, three model families, and a reasoning-distilled model. Qwen3-4B uses its original checkpoint on CONFLICTS and the Instruct-2507 release elsewhere, and Qwen3 runs without thinking. Relevance judgments and answers are greedy, except that R1 answers at temperature 0.1 and also uses it for relevance judgments on CONFLICTS and RAMDocs. Sampled judgments use temperature 0.7. On the three primary benchmarks and the controlled pools, every sampled Mistral-7B recipe is run three times, and a table reports the first run unless it states the mean of the three. All other sampled recipes run once. Unless stated otherwise, five documents fit in the answer context. Appendix~\ref{app:reproducibility} lists the checkpoints, token limits, and passage truncation.

\paragraph{Evaluation.}
We report answer-match accuracy (AM). After normalization, a prediction matches a gold answer when both strings are nonempty and one contains the other. RAMDocs follows its benchmark definition, under which a prediction must match every gold answer and contain no listed wrong answer~\citep{wang2025madamrag}. Since most RAMDocs questions have several gold answers, we also test an answer instruction that asks for every supported answer. With Mistral-7B, it raises the accuracy of \nf{}, \mpdrel{}, and \scscoring{} by one to three points, and \mpdrel{} remains the most accurate. For R1, we evaluate the text after the final reasoning delimiter. NQ Open uses 497 questions for Mistral-7B, Llama-3.1-70B, and R1, and 500 for Qwen3, as Appendix~\ref{app:reproducibility} details. We compare recipes on the same questions with paired bootstrap intervals over 10{,}000 resamples and exact McNemar tests, and Appendix~\ref{app:statistics} gives every paired effect. We call a difference reliable when its 95\% interval excludes zero. A pooled contrast averages the paired differences of the five models on a benchmark and resamples questions as clusters. As a check on the matching rule, exact match and token F1 preserve the sign of every difference from \nf{} on CONFLICTS and NQ Open.

\section{How Much Assessment Is Needed?}
\label{sec:results}
We first compare a single relevance judgment with unfiltered retrieval and with three quality judgments across models. We then locate the source of its gain and test whether additional calls add to it.
\subsection{One Relevance Judgment across Settings}

\begin{table*}[t]
\centering
\small
\setlength{\tabcolsep}{4pt}
\begin{tabular}{l>{\columncolor{white}}c>{\columncolor{methodblue}}cc>{\columncolor{white}}c>{\columncolor{methodblue}}cc>{\columncolor{white}}c>{\columncolor{methodblue}}cc}
\toprule
& \multicolumn{3}{c}{\textbf{CONFLICTS}} & \multicolumn{3}{c}{\textbf{RAMDocs}} & \multicolumn{3}{c}{\textbf{NQ Open}} \\
\cmidrule(lr){2-4}\cmidrule(lr){5-7}\cmidrule(lr){8-10}
\textbf{Assessor and answerer} & \nf{} & \mpdrel{} & \scscoring{} & \nf{} & \mpdrel{} & \scscoring{} & \nf{} & \mpdrel{} & \scscoring{} \\
\midrule
Mistral-7B & \textbf{65.8} & 63.7 & 62.9 & 18.2 & \textbf{20.6} & 19.2 & 32.6 & 33.8 & \textbf{35.6} \\
Llama-3.1-70B-AWQ & 69.6 & \textbf{76.4} & 75.1 & 21.6 & \textbf{22.8} & 22.4 & 40.6 & 42.7 & \textbf{42.9} \\
Qwen3-4B$^\dagger$ & 67.1 & \textbf{72.6} & 72.2 & 18.2 & \textbf{19.8} & 19.6 & 33.8 & 37.6 & \textbf{38.4} \\
Qwen3-8B & 70.0 & \textbf{73.8} & 71.3 & \textbf{14.6} & 14.2 & 14.4 & 36.0 & 38.0 & \textbf{38.8} \\
R1-Distill-Llama-8B & 54.4 & 57.0 & \textbf{59.5} & 14.0 & \textbf{14.4} & 14.2 & 29.4 & 33.8 & \textbf{35.6} \\
\bottomrule
\end{tabular}
\caption{Answer-match accuracy (\%) with five documents in the answer context. The same checkpoint assesses and answers in each row. Bold marks the most accurate recipe for each model and benchmark, and shading marks relevance scoring. $^\dagger$Instruct-2507 on RAMDocs and NQ Open.}
\label{tab:main_results}
\end{table*}

Table~\ref{tab:main_results} shows that assessment improves on unfiltered retrieval in most model and benchmark blocks, from 4B to 70B parameters and for the reasoning-distilled model. RAMDocs leaves less room because its released order already favors \nf{} and because the gain depends on how many documents fit in the context, as Section~\ref{sec:where} shows. Between the two assessed recipes, however, we find no consistent difference. Repeated quality scoring leads on NQ Open and trails in most CONFLICTS rows, and the paired 95\% interval of their difference includes zero in every row. Pooled over the five models, as Appendix~\ref{app:pooled} reports, relevance scoring improves on \nf{} with a 95\% interval above zero on every benchmark, and repeated quality scoring is not reliably more accurate than relevance scoring on CONFLICTS or RAMDocs. On NQ Open, repeated scoring leads by about one point, but the lead does not come from repeating the judgment, since a single quality judgment matches the accuracy of three for Llama-3.1-70B and exceeds it for Mistral-7B, as Appendix~\ref{app:k_sensitivity} shows. Because one relevance judgment improves on \nf{} on every benchmark with a third of the assessment calls, we use it as the reference below.

\subsection{Where the Gain Comes From}
\label{sec:where}

\begin{figure*}[t]
\centering
\includegraphics[width=\textwidth]{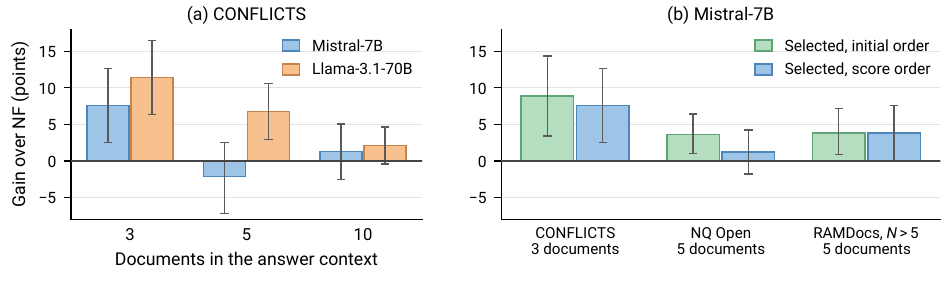}
\caption{Gain of relevance scoring over \nf{} in answer-match points, with paired 95\% bootstrap intervals. Panel (a) varies the number of documents in the answer context, and panel (b) sends the selected documents in initial or score order. $N$ counts pool candidates.}
\label{fig:budget_selection}
\end{figure*}

We find that the gain of relevance scoring comes from document selection, as Figure~\ref{fig:budget_selection} shows. In panel (a), the gain on CONFLICTS is largest when only three documents fit, reaching 7.6 points for Mistral-7B and 11.4 for Llama-3.1-70B. With ten documents, 233 of the 237 pools fit entirely, so assessment can only reorder them, and the gain falls to 1.3 points for Mistral-7B and 2.1 for Llama-3.1-70B.

Panel (b) separates selection from presentation order. In each setting of the panel, sending the selected documents in their initial order yields at least the gain of sending them in score order, and on NQ Open it yields more. More than half of the RAMDocs pools fit within five documents and leave only the order to change, so the panel keeps its 238 larger pools, where selection gains 3.8 points in either order. Pooled over the five models on NQ Open, relevance scoring still improves on \nf{} with a 95\% interval above zero when its selected documents keep their initial order. Deciding which documents enter the context is therefore the main source of the benefit.

The selection-only control also explains the lead of repeated scoring on NQ Open for Mistral-7B in Table~\ref{tab:main_results}. Sent in their initial order, the documents selected by \mpdrel{}, \scscoring{}, and \qualityonce{} reach 36.2, 36.4, and 36.4, although the recipes select different sets for about three quarters of the questions. The lead thus reflects the order of presentation and not a better selection. Finally, the gain does not depend on the released RAMDocs order. Shuffling the candidates of every pool lowers the accuracy of \nf{} by 1.4 points averaged over the five models, and relevance scoring then improves on \nf{} by 1.2 points with a 95\% interval above zero. Appendix~\ref{app:k_sensitivity} adds the RAMDocs budgets, and Appendix~\ref{app:pooled} the five-model order controls.

\subsection{Additional Judgments and Roles}

\begin{table}[t]
\centering
\small
\setlength{\tabcolsep}{4pt}
\begin{tabular}{lccc}
\toprule
\textbf{Recipe} & \textbf{Calls} & \textbf{CONFLICTS} & \textbf{RAMDocs} \\
\midrule
\nf{} & $1$ & 56.1 & 16.4 \\
\rowcolor{groupyellow}\multicolumn{4}{c}{\textit{One judgment for each document}} \\
\rowcolor{methodblue}\mpdrel{} & $N+1$ & 63.7 & \textbf{19.2} \\
\relonce{} & $N+1$ & 62.0 & 18.4 \\
\qualityonce{} & $N+1$ & 64.6 & 17.8 \\
\rowcolor{groupyellow}\multicolumn{4}{c}{\textit{Three judgments for each document}} \\
\screl{} & $3N+1$ & 62.4 & 18.4 \\
\scscoring{} & $3N+1$ & 64.1 & 18.4 \\
\rowcolor{groupyellow}\multicolumn{4}{c}{\textit{Additional roles or answers}} \\
\mpdthree{} & $2N+2$ & \textbf{65.8} & 18.6 \\
\scanswer{} & $3$ & 56.1 & 16.4 \\
\bottomrule
\end{tabular}
\caption{Mistral-7B answer-match accuracy (\%) with three documents in the answer context. \mpdrel{} and \mpdthree{} judge greedily, and the other assessed recipes and \scanswer{} sample at temperature 0.7. Calls include answer generation for $N$ candidates, and bold marks the best recipe in each column.}
\label{tab:role_ablation}
\end{table}

Table~\ref{tab:role_ablation} compares every recipe with three documents in the answer context, the budget at which selection gains most. We find that every recipe that reassesses the documents improves on \nf{}, whereas none that spends additional calls is reliably more accurate than \mpdrel{}. On CONFLICTS, \mpdthree{} is the most accurate with twice the assessment calls of \mpdrel{}, but on RAMDocs it trails \mpdrel{}, and the paired interval of their difference includes zero on both benchmarks.

The one-factor controls show that repetition itself has little effect on accuracy. Averaged over the three runs, three sampled judgments change the accuracy of a single sampled judgment by at most 0.6 points with either instruction, and \scanswer{}, which spends its calls without changing the evidence, falls slightly below \nf{}. One reason is that averaging three judgments mostly leaves the selected documents unchanged. In every run and at both budgets, three averaged judgments and one judgment of the same instruction select the same documents for at least two thirds of the questions and differ in correctness on at most 14 of the 237 CONFLICTS questions and 12 of the 500 RAMDocs questions. With five documents, the assessed recipes again stay within about two points of each other, and the three-run averages show no gain from averaging or voting. With Qwen3-8B on CONFLICTS, repeating either instruction, adding roles, or voting over answers likewise brings no reliable gain over the corresponding single pass at either budget. Appendix~\ref{app:k_sensitivity} lists all three runs at both budgets and the Qwen3-8B recipes.

\subsection{Controlled Distractors}

\begin{table}[t]
\centering\small
\setlength{\tabcolsep}{4pt}
\begin{tabular}{lccc}
\toprule
\textbf{Recipe} & \textbf{\shortstack{Assessment\\calls}} & \textbf{Generated} & \textbf{\shortstack{Other-\\question}} \\
\midrule
\nf{} & $0$ & 41.6 & 71.4 \\
\rowcolor{methodblue}\mpdrel{} & $N$ & 47.8 & \textbf{78.2} \\
\qualityonce{} & $N$ & \textbf{51.4} & 77.2 \\
\scscoring{} & $3N$ & 51.0 & 76.8 \\
\bottomrule
\end{tabular}
\caption{Mistral-7B answer-match accuracy (\%) on 500 TriviaQA questions whose pools add either generated distractors or passages of other questions. Bold marks the best recipe in each column.}
\label{tab:controlled_noise}
\end{table}

Table~\ref{tab:controlled_noise} tests whether these conclusions depend on the kind of distractor. The questions and their evidence stay fixed, and only the injected distractors vary. Every assessed recipe improves on \nf{} under both constructions, and Appendix~\ref{app:controlled_metrics} shows the same with exact match and token F1. Passages of other questions are easy to set apart from the evidence, since every assessed recipe raises the share of evidence among the selected documents from under half to about three quarters, and one relevance judgment is at least as accurate as the sampled recipes in all three runs. Generated passages that assert a plausible wrong answer are harder to set apart, but assessment still improves the answers. In both constructions, averaging three quality judgments adds no reliable gain to a single one.

\subsection{Single-Pass Rankers}

\begin{table}[t]
\centering\small
\setlength{\tabcolsep}{6pt}
\begin{tabular}{lccc}
\toprule
\textbf{Assessor} & \textbf{CONFLICTS} & \textbf{RAMDocs} & \textbf{NQ Open} \\
\midrule
\nf{} & 65.8 & 18.2 & 32.6 \\
\rowcolor{methodblue}\mpdrel{} & 63.7 & \textbf{20.6} & 33.8 \\
MiniLM & \textbf{68.4} & 18.4 & 35.2 \\
BGE & 67.9 & 20.4 & \textbf{36.2} \\
RankGPT & 66.2 & 17.0 & 35.8 \\
\bottomrule
\end{tabular}
\caption{Mistral-7B answer-match accuracy (\%) when single-pass assessors select the five documents for the same answer interface. MiniLM and BGE are cross-encoders, and RankGPT orders the whole pool in one call. Bold marks the best assessor in each column.}
\label{tab:rankers}
\end{table}

Table~\ref{tab:rankers} checks whether the single pass must be an LLM judgment by comparing \mpdrel{} with other single-pass assessors, whose configurations Appendix~\ref{app:rankgpt} describes. We observe that no assessor wins everywhere. MiniLM leads on CONFLICTS and BGE on NQ Open, where its gain over \nf{} has a paired interval above zero, and \mpdrel{} leads on RAMDocs, whose released order already favors \nf{}. Which single pass works best thus depends on the benchmark, and a cross-encoder that needs no LLM call is a competitive choice.

\section{Choosing a Recipe without Labels}
\label{sec:entropy}
A single pass is a strong default, but the preceding comparisons all required answer labels, which a new retrieval setting rarely provides. We therefore ask whether the relevance scores alone can indicate which recipe deserves validation first.
\subsection{A Score-Entropy Pilot}
For a pilot question $q$, let $p_q(v)$ be the fraction of its candidates with relevance score $v\in\{0,\ldots,5\}$. We compute
\begin{align}
 p_q(v)&=\frac{1}{|\mathcal D_q|}\sum_i\mathbf1[r_i=v],\\
 H(q)&=-\sum_{v=0}^{5}p_q(v)\log_2 p_q(v),
\end{align}
with $0\log_2 0=0$, and average $H(q)$ over the pilot questions to obtain $\bar H$. A low $\bar H$ means that the assessor places most candidates on a few score levels, so its scores separate them only coarsely. The pilot needs one scoring pass over a few pools and uses neither answer labels nor token probabilities.

\begin{figure}[t]
\centering
\includegraphics[width=\columnwidth]{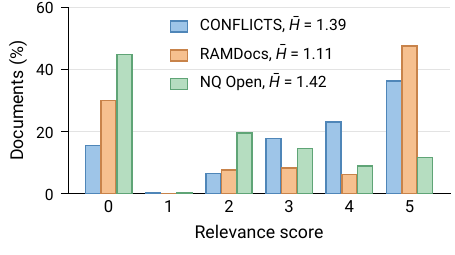}
\caption{Distribution of Mistral-7B relevance scores over all candidates of the 458 CONFLICTS, 500 RAMDocs, and 497 NQ Open pools. $\bar H$ averages the entropy of each question's score histogram.}
\label{fig:score_distributions}
\end{figure}

Figure~\ref{fig:score_distributions} illustrates what the entropy captures. CONFLICTS and NQ Open spread their relevance scores over several levels, whereas RAMDocs concentrates them at the two endpoints. As a result, more of its candidates share a score and its mean entropy is lower. The pilot turns this reading into a rule. When $\bar H<1.1$ and the candidate construction is known to contain entity or evidence ambiguity, it recommends validating \scscoring{} first, since this is the recipe of the main comparison that scores the candidates with another instruction. Otherwise, it recommends \mpdrel{}. A mean entropy of 1.1 bits lies just above the one bit of two equally used score levels, and the same threshold applies to every benchmark. All three primary benchmarks lie above it, so the rule keeps \mpdrel{}, which improves on \nf{} on each of them when pooled over the five models. The recommendation only decides which recipe to validate first. Once labels are available, a matched comparison that includes \nf{} makes the final choice and keeps the recipe with the fewest assessment calls whose accuracy is within one point of the best. Appendix~\ref{app:entropy_routing} states the full decision rule.

\subsection{Recommendations on Held-out Benchmarks}

\begin{table}[t]
\centering
\small
\setlength{\tabcolsep}{5pt}
\begin{tabular}{lcccc}
\toprule
\textbf{Benchmark} & $\bar H$ & \nf{} & \mpdrel{} & \scscoring{} \\
\midrule
MuSiQue & 1.08 & 11.6 & 24.0 & \textbf{25.6} \\
AmbigQA & 1.03 & 45.4 & \textbf{52.8} & 52.2 \\
\bottomrule
\end{tabular}
\caption{Mistral-7B answer-match accuracy (\%) on the 500 questions of each held-out benchmark. $\bar H$ is the mean score entropy of the 50-question pilot, and bold marks the most accurate recipe.}
\label{tab:heldout}
\end{table}

We test the rule on the two held-out benchmarks in Table~\ref{tab:heldout}. Both pilot entropies fall below the threshold and both constructions are known to contain ambiguity. The pilot therefore recommends \scscoring{} for both. The mean entropy over all 500 evaluated pools, 1.00 on MuSiQue and 1.09 on AmbigQA, gives the same recommendation. On MuSiQue, the recommended recipe is the most accurate, and both assessed recipes roughly double the accuracy of \nf{}. On AmbigQA, the two assessed recipes are nearly tied, and the labeled comparison keeps \mpdrel{}, which is slightly more accurate and needs fewer calls. On both benchmarks, the pilot's recommendation is the most accurate recipe or within a point of it. Moreover, a single quality judgment, which keeps the instruction of the recommended recipe without repeating it, reaches 24.4 on MuSiQue and 53.0 on AmbigQA, so repeating the judgment adds no reliable gain on either benchmark. Appendix~\ref{app:entropy_routing} adds exact match, token F1, and the paired effects of these comparisons.

The pilot characterizes one assessor on one candidate-pool construction, so a new assessor or pool construction needs a new pilot. A new answer generator or context budget leaves the relevance scores unchanged and needs only a new labeled comparison.

\section{Conclusion}
We asked how much document assessment a RAG system with a fixed retriever and answer generator needs, and under a matched protocol the answer is a single pass over the candidates. Pooled over five models, one relevance judgment for each document improves on unfiltered retrieval on every benchmark. Further judgments, assessor roles, or answer votes add no reliable gain, and where three quality judgments lead slightly, the lead does not come from repeating the judgment. The gain comes from which documents enter a limited context and tends to be largest when few documents fit, whereas sending the chosen documents in score order is not reliably better than keeping their initial order. For a new retrieval setting, an unlabeled score-entropy pilot can suggest which recipe to validate first, and on two held-out benchmarks it recommends the most accurate recipe or one nearly tied with it. We therefore recommend starting from a single assessment pass and spending further calls only where a matched comparison shows that they change the answers.

\section*{Limitations}
Our comparisons cover English short-answer question answering with open-weight models of up to 70B parameters and with candidate pools of at most twenty passages fixed before assessment. Long-form generation would additionally require measuring how completely the selected evidence covers the answer. In addition, the costlier recipes isolate repetition, extra roles, and answer voting within one shared interface, which leaves complete multi-agent systems with debate or iterative retrieval outside the comparison. Likewise, the same checkpoint assesses and answers in every comparison, and pairing a small assessor with a larger generator may shift the balance between assessment cost and accuracy. We measure inference work in model calls, while token counts and latency also depend on input lengths, batching, and hardware. Finally, the score-entropy pilot uses a six-level relevance rubric and recommends one recipe for a whole retrieval setting, so its threshold should be recalibrated for a different rubric or assessor.

\section*{Ethics Statement}
We use public question-answering datasets and open model checkpoints under their release terms. The misleading passages in our controlled comparisons serve only as evaluation material within fixed candidate pools and are not presented as factual sources. Document assessment can reduce exposure to irrelevant or misleading evidence, but it does not by itself make a generated answer reliable~\citep{lee2026layermix}.

\bibliographystyle{acl_natbib}

\appendix

\section{Data and Reproduction Details}
\label{app:reproducibility}
\label{app:eval_protocol}
\label{app:cross_model}

\paragraph{CONFLICTS and RAMDocs.}
CONFLICTS supplies pools for 458 questions with 5 to 20 search results each, 9.2 on average. Its authors collected the results with Google Search and reduced each page to its most question-relevant window of 512 tokens. Answer accuracy uses the 237 questions with single-answer annotations, while the score histograms use all 458 pools. When a publication date is supplied, it prefixes the passage text for assessment and follows the document marker in the answer context. RAMDocs supplies 500 pools with 1 to 12 documents labeled as correct, misinformation, or noise, and 262 of them contain at most five documents. Its released order lists a correct document first in 476 pools and places noise documents mostly at the end. In addition, 400 of its questions have more than one gold answer. We keep the released pools of both benchmarks without new retrieval.

\paragraph{NQ Open retrieval.}
Question text queries English Wikipedia search. Up to five retrieved pages are split at sentence boundaries into passages of approximately 200 words, discarding passages shorter than 20 words. A lexical scorer retains up to ten passages using the BM25 term-frequency and length-normalization formula with $k_1=1.5$, $b=0.75$, and every inverse-document-frequency weight fixed to one. Average passage length is computed within the question's page-derived candidate set. Gold aliases are used afterward to annotate whether a retained passage contains an answer and never enter retrieval or ranking.

The Mistral-7B and Llama-3.1-70B comparisons use the same 497 questions and ordered pools, drawn from a 500-question validation sample after requiring at least five retrieved passages. The two Qwen3 comparisons share another 500 validation questions, and R1 uses the first 497 of them in the same order. The two question sets share 379 questions, whose candidate texts and order agree. NQ Open comparisons are therefore paired within each model row.

\paragraph{MuSiQue and AmbigQA.}
MuSiQue uses the first 500 eligible questions in the supplied data order, each with answer annotations and at least five nonempty supplied passages. The assessors receive the question and passage text, without gold-answer or support labels.

AmbigQA uses 500 questions from the full validation split with at least two distinct annotated answer strings. Each ten-document pool takes up to three passages from its annotated Wikipedia page, followed by one top passage from each other page in question-text search with up to eight hits. Passages are ranked by question-term occurrences divided by the square root of passage length, using normalized tokens. Remaining slots use random-page passages when available, then the fixed text ``This document does not contain relevant information.'' All ten positions are shuffled. Fixed padding occupies 1{,}484 of the 5{,}000 slots. The task is to return one acceptable listed answer, not to enumerate every interpretation.

\paragraph{Models and decoding.}
Table~\ref{tab:primary_protocol} lists the checkpoints used for both assessment and answering. Qwen3 uses the non-thinking setting. R1 answering uses temperature 0.1 and repetition penalty 1.05. Its CONFLICTS and RAMDocs relevance calls use the same temperature and penalty, whereas its NQ relevance call uses temperature 0 and penalty 1. Other primary relevance and answer calls use temperature 0. Repeated quality and relevance scoring use three samples at temperature 0.7, and the three Mistral-7B runs of a sampled recipe draw their samples independently. All assessment and answer calls run in vLLM~\citep{kwon2023vllm} and format their messages with the Hugging Face tokenizer~\citep{wolf2020transformers} and chat template of the checkpoint. The 70B checkpoint is \path{hugging-quants/Meta-Llama-3.1-70B-Instruct-AWQ-INT4}, with revision \texttt{2123003760781134cfc31\allowbreak124aa6560a45b491fdf}. Its quantization uses four bits, group size 128, zero points, and GEMM kernels.

\begin{table*}[t]
\centering\small
\setlength{\tabcolsep}{5pt}
\begin{tabular}{llcc}
\toprule
\textbf{Checkpoint used for both roles} & \textbf{Benchmarks} & \textbf{Score tokens} & \textbf{Answer tokens} \\
\midrule
Mistral-7B-Instruct-v0.3 & CONFLICTS, RAMDocs, NQ Open & 512 & 100 \\
Meta-Llama-3.1-70B-Instruct-AWQ-INT4 & CONFLICTS, RAMDocs, NQ Open & 512 & 100 \\
Qwen3-4B & CONFLICTS & 2048 & 512 \\
Qwen3-4B-Instruct-2507 & RAMDocs, NQ Open & 2048 & 512 \\
Qwen3-8B & CONFLICTS, RAMDocs, NQ Open & 2048 & 512 \\
DeepSeek-R1-Distill-Llama-8B & CONFLICTS, RAMDocs & 1024 & 512 \\
DeepSeek-R1-Distill-Llama-8B & NQ Open & 2048 & 512 \\
\bottomrule
\end{tabular}
\caption{Checkpoints and generated-token limits for the fifteen primary comparison blocks.}
\label{tab:primary_protocol}
\end{table*}

\paragraph{Input limits and score parsing.}
Relevance and quality assessment read at most 2{,}000 characters from each passage. Primary answer contexts retain at most five passages with 1{,}500 characters each. The held-out answer contexts keep up to 2{,}000 characters of each selected passage, marked as \texttt{Document j} and, on MuSiQue, followed by the page title. The consistency assessor reads one truncated document and the relevance evidence of the other documents. The joint conflict call receives those evidence summaries and relevance scores. The held-out comparisons of \nf{}, \mpdrel{}, and \scscoring{} and every Llama-3.1-70B run except the three- and ten-document budgets use a 4{,}096-token context limit, and all other runs use 8{,}192 tokens. Every held-out answer prompt holds at most 2{,}780 tokens and every ten-document prompt of Llama-3.1-70B fewer than 3{,}700, so neither limit truncates them.

The parser tries the full response, a fenced JSON block, and an embedded JSON object. If none of them yields a score and the response is a JSON object that lacks only its closing brace, the parser restores the brace. Parsed scores are converted to integers and clipped to $[0,5]$. A missing score field or unparsed response receives zero, which remains in both rankings and histograms. With Mistral-7B, this applies to fewer than 0.3\% of the greedy relevance responses on each primary benchmark. Initial candidate order breaks score ties.

\paragraph{Answer matching.}
Normalization lowercases text, removes punctuation and articles, and collapses whitespace. A gold match requires nonempty normalized strings and accepts either substring direction. CONFLICTS, NQ Open, TriviaQA, MuSiQue, and AmbigQA accept any gold alias. RAMDocs requires every gold answer to match and rejects any prediction containing a nonempty listed wrong answer. A question whose gold aliases all normalize to empty is incorrect under every recipe and remains in the denominator. For R1, evaluation uses the text after the last literal \texttt{</think>} delimiter, and an absent delimiter or empty final segment counts as incorrect.

\subsection{Controlled Distractor Construction}
\label{app:controlled_noise}

The controlled comparison uses the first 500 questions from the TriviaQA reading-comprehension validation subset. The first 1{,}000 questions supply the pool of other-question passages. For each question, we extract nonempty search snippets followed by Wikipedia-page contexts and retain up to five source passages. Five injected passages produce pools of six to ten documents, with a mean of 9.33. Each passage is truncated to 1{,}000 characters before assessment and answering.

The generated-distractor condition uses five passages prepared with Mistral-7B-Instruct-v0.3 at temperature 0.9 and a 200-token output limit. Figure~\ref{fig:prompt_distractor} gives the generation messages. Each request supplies the question and an answer alias and asks for a plausible incorrect alternative. The other-question condition samples five passages from the first three available evidence passages of other questions, excluding the current question without any entity, topic, or answer restriction. The conditions share questions and source texts, with separately shuffled candidate positions.

\begin{figure}[t]
\centering
\begin{Verbatim}[frame=single, rulecolor=\color{black!55}, framesep=4pt, breaklines=true, breakanywhere=false, breaksymbolleft={}, fontsize=\scriptsize, xleftmargin=2pt, xrightmargin=2pt]
System:
You are a creative writing assistant.

User:
Generate a SHORT fake document snippet (2-3 sentences) that answers the following question with WRONG information. The document should look realistic and authoritative but contain a factually incorrect answer.

Question: {question}
Correct answer (DO NOT use this): {answer}

Write ONLY the fake document text. Make it look like a real encyclopedia or news article excerpt. Use a different but plausible-sounding answer.
\end{Verbatim}
\caption{Messages used to generate the controlled distractors. The question and one gold alias are substituted before generation.}
\label{fig:prompt_distractor}
\end{figure}

Scoring and answer generation use Mistral-7B-Instruct-v0.3 in float16 with an 8{,}192-token context window and vLLM 0.15.1 on an H100 GPU. The scoring limit is 512 output tokens, and answers are greedy with a 100-token limit. All four recipes retain five passages and use the same answer messages.

\section{Selection Procedure and Instructions}
\label{app:algorithm}
\label{app:prompts}
\label{app:relevance_prompt}
\label{app:conflict_prompt}
\label{app:sc_scoring_prompt}
\label{app:token_cost}

Algorithm~\ref{alg:mpdrel} specifies the relevance-scoring branch. The assessed branches present selected documents in descending score order, whereas \nf{} preserves the initial candidate order. The selection-only control sends the documents selected by relevance scoring in increasing candidate position. In the Mistral-7B comparisons and in the controls with Qwen3-8B and with all five models, conditions that send the same documents in the same order share one generated answer. \relonce{} and \qualityonce{} keep the first of the three sampled judgments of \screl{} and \scscoring{}, respectively, whereas on MuSiQue and AmbigQA \qualityonce{} draws its own sampled judgment.

\begin{algorithm}[t]
\caption{Relevance scoring and fixed-budget selection.}
\label{alg:mpdrel}
\begin{algorithmic}[1]
\REQUIRE Question $q$, ordered documents $\mathcal D$, budget $k$
\FOR{each $d_i\in\mathcal D$}
\STATE $s_i\leftarrow\textsc{RelevanceScore}(q,d_i)$
\ENDFOR
\STATE Rank documents by decreasing $s_i$, breaking ties by input order
\STATE $\mathcal D'\leftarrow$ the first $\min(k,|\mathcal D|)$ ranked documents
\STATE $a\leftarrow\textsc{Answer}(q,\mathcal D')$
\RETURN $a$
\end{algorithmic}
\end{algorithm}

\paragraph{Instructions.}
Table~\ref{tab:instructions} summarizes every model call. Each Mistral call supplies a system instruction and a user message through the checkpoint's chat template. The user message states the question, followed by the passage or passages that the call reads, and ends with a one-line request for the output. Assessment outputs are JSON objects, and only their score fields enter selection. The relevance evidence field serves as the claim summary of the role recipe, without a separate extraction call.

\begin{table*}[t]
\centering\small
\setlength{\tabcolsep}{4pt}
\begin{tabular}{>{\raggedright\arraybackslash}p{0.11\textwidth}>{\raggedright\arraybackslash}p{0.22\textwidth}>{\raggedright\arraybackslash}p{0.58\textwidth}}
\toprule
\textbf{Call} & \textbf{Input and output} & \textbf{Instruction content} \\
\midrule
Answer & Question and selected passages; a short answer & Answer from the provided documents only, with a name, number, date, or short phrase and no explanation. Two short lists illustrate acceptable and unacceptable answer forms. \\
\addlinespace
Relevance & Question and one passage; score, evidence, rationale & Judge whether the passage offers evidence that can directly help answer the question, considering the question's entities and relations and separating real evidence from shared keywords. The levels run from 5, direct support for the answer, through strong support, partial or indirect support, keyword overlap only, and tangential relation, to 0 for an irrelevant passage. \\
\addlinespace
Quality & Question and one passage; score and rationale & Judge how useful the passage is for answering the question on the same six levels as the relevance instruction. \\
\addlinespace
Consistency & Question, one passage, and the evidence of the other passages; score and two assessments & Judge whether the passage is internally coherent, agrees with most other passages, and makes specific claims, from 5 for full consistency to 0 for contradiction. \\
\addlinespace
Conflict & Question and the evidence of all passages with relevance scores; conflict flag, conflicting pairs, majority position, adjustments & Identify contradictory answers and outdated information, and assign every passage an adjustment from $-2$ to $+2$ that rewards support from the majority and recent information. \\
\addlinespace
Listwise ranking & Question and all numbered passages; a ranking & The single-window instruction of RankGPT~\citep{sun2023rankgpt}, which asks for passage identifiers in descending relevance. \\
\bottomrule
\end{tabular}
\caption{Model calls used by the assessment recipes, with their inputs, outputs, and instruction content.}
\label{tab:instructions}
\label{tab:prompt_interfaces}
\end{table*}

\paragraph{Role scores.}
\mpdthree{} lists the other candidates as \texttt{Doc j: evidence} in the consistency input and all candidates as \texttt{Doc j (relevance=r/5): evidence} in the joint conflict input, using zero-based identifiers. Missing evidence is represented by a placeholder. The parser accepts adjustments that carry a plus sign or name their document as \texttt{Doc j}, and it keeps the adjustment object even when other fields of the response are malformed. The adjustments are converted to integers and clipped to $[-2,2]$, and candidates without an adjustment receive zero. With relevance $r_i$, consistency $u_i$, and conflict adjustment $a_i$, the role score is $0.5r_i+0.3u_i+0.2c_i$, where
\begin{equation}
c_i=\operatorname{clip}\Bigl(\frac{r_i+u_i}{2}+a_i,0,5\Bigr).
\end{equation}

\paragraph{Answer context and voting.}
For CONFLICTS and RAMDocs, the answer context joins selected passages with two newlines and prefixes each with \texttt{[Document j]:} for its position in the selected list. CONFLICTS adds the supplied date after the marker. The NQ Open contexts of Mistral-7B and Llama-3.1-70B also include document titles. No context includes assessor scores or explanations. \scanswer{} samples three answers at temperature 0.7 from the first three or five candidates in their initial order, matching the document budget. Voting selects the most frequent normalized answer, ties follow generation order, and the returned answer is the earliest original string in the winning class.

\section{Additional Analyses}
\label{app:k_sensitivity}
\label{app:rankgpt}
\label{app:crossencoder}
\label{app:scoring_ablation}
\label{app:error_analysis}
\label{app:full_results}
\label{app:factorial}

\paragraph{Document budget.}
Table~\ref{tab:k_sensitivity} lists every measured budget. On RAMDocs, the gain with Mistral-7B is also largest at three documents. We do not report a ten-document RAMDocs condition, because 238 of its pools exceed five documents but only seven exceed ten.

\begin{table}[t]
\centering\small
\setlength{\tabcolsep}{3pt}
\begin{tabular}{llcccc}
\toprule
\textbf{Benchmark} & \textbf{Model} & $k$ & \nf{} & \mpdrel{} & $\Delta$ \\
\midrule
CONFLICTS & Mistral-7B & 3 & 56.1 & 63.7 & +7.6 \\
 & & 5 & 65.8 & 63.7 & $-2.1$ \\
 & & 10 & 69.2 & 70.5 & +1.3 \\
 & Llama-3.1-70B & 3 & 59.9 & 71.3 & +11.4 \\
 & & 5 & 69.6 & 76.4 & +6.8 \\
 & & 10 & 74.7 & 76.8 & +2.1 \\
 & Qwen3-8B & 3 & 58.6 & 72.2 & +13.5 \\
 & & 5 & 70.0 & 73.8 & +3.8 \\
\midrule
RAMDocs & Mistral-7B & 3 & 16.4 & 19.2 & +2.8 \\
 & & 5 & 18.2 & 20.6 & +2.4 \\
 & Llama-3.1-70B & 3 & 20.2 & 21.4 & +1.2 \\
 & & 5 & 21.6 & 22.8 & +1.2 \\
\bottomrule
\end{tabular}
\caption{Answer-match accuracy (\%) of relevance scoring and \nf{} for every measured document budget $k$, with differences in points.}
\label{tab:k_sensitivity}
\end{table}

\paragraph{Repeated runs.}
Tables~\ref{tab:reruns} and~\ref{tab:reruns_other} summarize the three runs of every sampled Mistral-7B recipe beside the single runs of the greedy recipes. Averaged over the runs, three sampled judgments differ from one judgment of the same instruction by $-1.1$ to $+0.6$ points in every setting, and after each question's correctness is averaged over the runs, none of these differences has a paired interval above zero, while the interval of repeated quality scoring on RAMDocs with five documents lies below zero. Answer voting stays 0.3 to 0.7 points below \nf{} on average. On NQ Open, one quality judgment is more accurate than \mpdrel{} in every run, and Table~\ref{tab:selection_order} shows that the two selections reach the same accuracy in their initial order. With Llama-3.1-70B on NQ Open, \qualityonce{} reaches 42.7 and \scscoring{} 42.9, and the paired interval of their difference includes zero in either order.

\begin{table*}[t]
\centering\small
\setlength{\tabcolsep}{6pt}
\begin{tabular}{lcccc}
\toprule
& \multicolumn{2}{c}{\textbf{Three documents}} & \multicolumn{2}{c}{\textbf{Five documents}} \\
\cmidrule(lr){2-3}\cmidrule(lr){4-5}
\textbf{Recipe} & \textbf{CONFLICTS} & \textbf{RAMDocs} & \textbf{CONFLICTS} & \textbf{RAMDocs} \\
\midrule
\rowcolor{groupyellow}\multicolumn{5}{c}{\textit{Greedy recipes, one run}} \\
\nf{} & 56.1 & 16.4 & 65.8 & 18.2 \\
\mpdrel{} & 63.7 & 19.2 & 63.7 & 20.6 \\
\mpdthree{} & 65.8 & 18.6 & 64.1 & 20.4 \\
\rowcolor{groupyellow}\multicolumn{5}{c}{\textit{Sampled recipes, mean of three runs with the lowest and highest run}} \\
\relonce{} & 63.6 (62.0--64.6) & 18.9 (18.4--19.2) & 64.4 (63.7--65.0) & 20.5 (19.2--21.2) \\
\qualityonce{} & 63.7 (62.9--64.6) & 18.7 (17.8--19.6) & 64.0 (63.7--64.1) & 19.9 (19.6--20.0) \\
\screl{} & 63.4 (62.4--64.1) & 18.9 (18.4--19.2) & 64.3 (63.7--64.6) & 19.8 (18.8--20.4) \\
\scscoring{} & 64.3 (63.7--65.0) & 18.2 (18.0--18.4) & 63.9 (62.9--64.6) & 18.9 (18.6--19.2) \\
\scanswer{} & 55.7 (55.3--56.1) & 16.1 (15.4--16.4) & 65.5 (65.0--66.7) & 17.5 (16.8--18.6) \\
\bottomrule
\end{tabular}
\caption{Mistral-7B answer-match accuracy (\%) on CONFLICTS and RAMDocs at both document budgets, with each sampled recipe run three times.}
\label{tab:reruns}
\end{table*}

\begin{table}[t]
\centering\small
\setlength{\tabcolsep}{2.4pt}
\begin{tabular}{lccc}
\toprule
\textbf{Recipe} & \textbf{NQ Open} & \textbf{Generated} & \textbf{\shortstack{Other-\\question}} \\
\midrule
\nf{} & 32.6 & 41.6 & 71.4 \\
\mpdrel{} & 33.8 & 47.8 & 78.2 \\
\qualityonce{} & 37.0 {\scriptsize(35.8--38.0)} & 51.5 {\scriptsize(50.4--52.6)} & 77.7 {\scriptsize(77.2--78.2)} \\
\scscoring{} & 35.9 {\scriptsize(35.6--36.0)} & 51.3 {\scriptsize(50.6--52.4)} & 77.3 {\scriptsize(76.8--78.0)} \\
\bottomrule
\end{tabular}
\caption{Mistral-7B answer-match accuracy (\%) with five documents on NQ Open and the controlled pools. Sampled recipes show the mean of three runs with the lowest and highest run.}
\label{tab:reruns_other}
\end{table}

\paragraph{Controls with Qwen3-8B.}
Table~\ref{tab:qwen_controls} reports every recipe with Qwen3-8B on CONFLICTS, with one run of each sampled recipe drawn independently of the run in Table~\ref{tab:main_results}. With three documents, every recipe that reassesses the documents improves on \nf{} by 11.8 to 13.5 points with paired intervals above zero. At both budgets, the paired intervals of \screl{} against \relonce{}, \scscoring{} against \qualityonce{}, \mpdthree{} against \mpdrel{}, and \scanswer{} against \nf{} include zero.

\begin{table}[t]
\centering\small
\setlength{\tabcolsep}{5pt}
\begin{tabular}{lccc}
\toprule
\textbf{Recipe} & \textbf{Calls} & $k=3$ & $k=5$ \\
\midrule
\nf{} & $1$ & 58.6 & 70.0 \\
\rowcolor{groupyellow}\multicolumn{4}{c}{\textit{One judgment for each document}} \\
\rowcolor{methodblue}\mpdrel{} & $N+1$ & \textbf{72.2} & 73.8 \\
\relonce{} & $N+1$ & 71.3 & \textbf{74.3} \\
\qualityonce{} & $N+1$ & 70.9 & 71.3 \\
\rowcolor{groupyellow}\multicolumn{4}{c}{\textit{Three judgments for each document}} \\
\screl{} & $3N+1$ & 71.7 & \textbf{74.3} \\
\scscoring{} & $3N+1$ & 70.5 & 70.9 \\
\rowcolor{groupyellow}\multicolumn{4}{c}{\textit{Additional roles or answers}} \\
\mpdthree{} & $2N+2$ & 70.9 & 71.7 \\
\scanswer{} & $3$ & 58.2 & 70.0 \\
\bottomrule
\end{tabular}
\caption{Qwen3-8B answer-match accuracy (\%) on CONFLICTS with three or five documents in the answer context. Calls include answer generation for $N$ candidates, and bold marks the best recipe in each column.}
\label{tab:qwen_controls}
\end{table}

\paragraph{Selection and order.}
Table~\ref{tab:selection_order} gives the accuracies behind the order comparison.

\begin{table}[t]
\centering\small
\setlength{\tabcolsep}{4pt}
\begin{tabular}{lcccc}
\toprule
\textbf{Setting} & $n$ & \nf{} & \textbf{\shortstack{Initial\\order}} & \textbf{\shortstack{Score\\order}} \\
\midrule
\rowcolor{groupyellow}\multicolumn{5}{c}{\textit{Selected by \mpdrel{}}} \\
CONFLICTS, $k=3$ & 237 & 56.1 & 65.0 & 63.7 \\
CONFLICTS, $k=5$ & 237 & 65.8 & 65.4 & 63.7 \\
NQ Open & 497 & 32.6 & 36.2 & 33.8 \\
RAMDocs & 500 & 18.2 & 20.0 & 20.6 \\
RAMDocs, $N>5$ & 238 & 9.2 & 13.0 & 13.0 \\
\rowcolor{groupyellow}\multicolumn{5}{c}{\textit{NQ Open, selected by a sampled recipe}} \\
\scscoring{} & 497 & 32.6 & 36.4 & 35.6 \\
\qualityonce{} & 497 & 32.6 & 36.4 & 38.0 \\
\bottomrule
\end{tabular}
\caption{Mistral-7B answer-match accuracy (\%) when the selected documents are sent in their initial order or in score order.}
\label{tab:selection_order}
\end{table}

\paragraph{Role-score combinations.}
Table~\ref{tab:scoring_ablation} recombines the RAMDocs relevance, consistency, and conflict scores at budget five without regenerating answers. Here $c_a=\operatorname{clip}((r+u)/2+a,0,5)$ is the conflict score of \mpdthree{}. Precision is the relevant share of the selected set, and recall is the retained share of relevant candidates, averaged across questions. Recall omits the three questions without a relevant candidate. Every combination selects evidence of nearly the same precision and recall as relevance alone, which matches the small accuracy differences between \mpdthree{} and \mpdrel{}.

\begin{table*}[t]
\centering\small
\setlength{\tabcolsep}{5pt}
\begin{tabular}{llcccc}
\toprule
\textbf{Rule} & \textbf{Formula} & \textbf{Precision} & \textbf{Recall} & \textbf{Misinfo.} & \textbf{Noise} \\
\midrule
Neutral-centered conflict & $0.5r+0.3u+0.2(2.5+a)$ & 73.4 & 90.1 & 0.48 & 0.65 \\
Mean-centered conflict & $0.5r+0.3u+0.2c_a$ & 73.4 & 90.1 & 0.48 & 0.65 \\
\rowcolor{methodblue}Relevance & $r$ & 72.7 & 89.5 & 0.54 & 0.62 \\
Relevance and consistency & $0.6r+0.4u$ & 73.1 & 89.8 & 0.50 & 0.64 \\
Equal anchored weights & $(r+u+c_a)/3$ & 73.6 & 90.2 & 0.47 & 0.65 \\
\bottomrule
\end{tabular}
\caption{Selection-only score combinations on 500 RAMDocs questions at budget five. Precision and recall are percentages, and misinformation and noise are mean counts of selected documents.}
\label{tab:scoring_ablation}
\end{table*}

\paragraph{Single-pass rankers.}
All rankers select five documents for the primary Mistral-7B answer interface. The cross-encoders \href{https://huggingface.co/cross-encoder/ms-marco-MiniLM-L-12-v2}{ms-marco-MiniLM-L-12-v2}~\citep{wang2020minilm}, with 33 million parameters, and BGE-reranker-v2-m3~\citep{chen2024bge}, with 568 million, read the same passage text as the relevance assessor, truncated to 512 tokens, and send their five highest-scoring passages in score order. RankGPT~\citep{sun2023rankgpt} uses Mistral-7B with greedy decoding and a 200-token output limit. Its single window contains every candidate in the initial order, using the first 500 characters of each passage and the date when supplied, and its first five ranked passages are sent in rank order. The parser keeps the first occurrence of each valid identifier and appends omitted candidates in their initial order. Paired effects of all rankers are listed in Table~\ref{tab:stats_rankers}.

\section{Score-Entropy Pilot}
\label{app:entropy_routing}
\label{app:evidence_hierarchy}
\label{app:score_distributions}
\label{app:entropy_sensitivity}
\label{app:heldout}
\label{app:ambigqa_prospective}

\paragraph{Initial recommendation and labeled selection.}
The pilot computes the six-bin relevance-score entropy within each question's candidate pool and averages it over the pilot questions. Let $A$ indicate entity or evidence ambiguity known from the candidate construction. The initial recommendation is
\begin{equation}
\hat m=\begin{cases}
\text{\scscoring{}}, & \bar H<1.1\ \text{and}\ A,\\
\text{\mpdrel{}}, & \text{otherwise}.
\end{cases}
\end{equation}
Low entropy without known ambiguity keeps relevance scoring, as does entropy at or above the threshold. If labels are available, the final comparison includes \nf{} and chooses the recipe with the fewest assessment calls whose AM is within one point of the best.

\paragraph{Pilot sampling.}
MuSiQue and AmbigQA each use the first 50 of their 500 questions as the unlabeled pilot, and these questions remain in the answer evaluation. MuSiQue keeps its data-file order. AmbigQA keeps questions with at least two distinct annotated answer strings, sorts them by that count, shuffles up to the first 1{,}000 eligible questions, and retains 500.

The pilot entropies are 1.0763 bits for MuSiQue and 1.0263 bits for AmbigQA. Both recommend repeated quality scoring, since partial evidence chains and related answer entities provide the ambiguity indicator of the two constructions. Over all 500 evaluated pools, the mean entropy is 1.0008 bits for MuSiQue and 1.0946 bits for AmbigQA, so the full question sets lead to the same recommendation. The fixed padding of AmbigQA always receives a relevance score of zero, and without it the AmbigQA pilot entropy is 1.0273 bits.

\paragraph{Held-out answer metrics.}
Table~\ref{tab:prospective_standard_metrics} evaluates the two held-out benchmarks with EM and token F1, together with one sampled quality judgment. All three assessed recipes improve over \nf{} under every metric and stay within two points of one another. A padding-skip rule, which sends the first five non-padding candidates in their initial order and fills any remaining slots with padding, reaches 47.4 AM, 28.0 EM, and 41.9 F1, below both \mpdrel{} and \scscoring{}. AmbigQA scores the best match among its listed answer alternatives.

\begin{table}[t]
\centering\small
\setlength{\tabcolsep}{3.5pt}
\begin{tabular}{lcccccc}
\toprule
& \multicolumn{3}{c}{\textbf{MuSiQue}} & \multicolumn{3}{c}{\textbf{AmbigQA}} \\
\cmidrule(lr){2-4}\cmidrule(lr){5-7}
\textbf{Recipe} & \textbf{AM} & \textbf{EM} & \textbf{F1} & \textbf{AM} & \textbf{EM} & \textbf{F1} \\
\midrule
\nf{} & 11.6 & 3.8 & 10.9 & 45.4 & 25.8 & 39.7 \\
\mpdrel{} & 24.0 & 12.2 & 22.3 & 52.8 & 32.6 & 47.6 \\
\scscoring{} & 25.6 & 11.2 & 22.8 & 52.2 & 31.6 & 46.0 \\
\qualityonce{} & 24.4 & 11.6 & 21.8 & 53.0 & 32.2 & 47.2 \\
\bottomrule
\end{tabular}
\caption{Answer metrics (\%) with Mistral-7B on the 500 questions of each held-out benchmark.}
\label{tab:prospective_standard_metrics}
\end{table}

\paragraph{Assessor dependence.}
Changing the CONFLICTS assessor from Mistral-7B to Llama-3.1-70B shifts its mean entropy from 1.39 to 1.28 bits and increases the share of score-5 judgments. A new assessor or candidate pool therefore calls for a new pilot, whereas a new answer generator can leave the scores unchanged while altering the accuracy comparison.

\section{Paired Effects and Uncertainty}
\label{app:statistics}
\label{app:main_ci}
\label{app:forest}
\label{app:controlled_metrics}

\subsection{Conventional Answer Metrics}
Table~\ref{tab:standard_metrics} evaluates the primary CONFLICTS and NQ Open answers with EM and token F1. EM requires nonempty normalized equality with a gold alias. Token F1 is the harmonic mean of multiset token precision and recall, taking the maximum over gold aliases and then averaging across questions. Both metrics use the same normalization and final-answer extraction as AM, and both preserve the sign of every AM difference from \nf{} in Table~\ref{tab:main_results}.

\begin{table*}[t]
\centering\small
\setlength{\tabcolsep}{6pt}
\begin{tabular}{lccccccc}
\toprule
& & \multicolumn{2}{c}{\nf{}} & \multicolumn{2}{c}{\mpdrel{}} & \multicolumn{2}{c}{\scscoring{}} \\
\cmidrule(lr){3-4}\cmidrule(lr){5-6}\cmidrule(lr){7-8}
\textbf{Model} & $n$ & \textbf{EM} & \textbf{F1} & \textbf{EM} & \textbf{F1} & \textbf{EM} & \textbf{F1} \\
\midrule
\rowcolor{groupyellow}\multicolumn{8}{c}{\textit{CONFLICTS}} \\
Mistral-7B & 237 & 47.3 & 59.7 & 45.1 & 57.4 & 46.0 & 57.6 \\
Llama-3.1-70B-AWQ & 237 & 62.0 & 67.8 & 67.9 & 73.6 & 66.2 & 71.6 \\
Qwen3-4B & 237 & 58.2 & 65.4 & 62.0 & 69.6 & 62.0 & 69.3 \\
Qwen3-8B & 237 & 59.9 & 66.8 & 63.3 & 70.0 & 61.6 & 68.2 \\
R1-Distill-Llama-8B & 237 & 37.1 & 44.8 & 41.8 & 49.8 & 44.3 & 51.8 \\
\rowcolor{groupyellow}\multicolumn{8}{c}{\textit{NQ Open}} \\
Mistral-7B & 497 & 17.5 & 28.4 & 18.3 & 29.5 & 20.1 & 30.6 \\
Llama-3.1-70B-AWQ & 497 & 31.8 & 40.5 & 33.2 & 42.1 & 33.8 & 42.5 \\
Qwen3-4B-Instruct-2507 & 500 & 22.4 & 31.3 & 24.4 & 33.9 & 24.6 & 34.5 \\
Qwen3-8B & 500 & 24.4 & 34.8 & 25.8 & 36.3 & 26.4 & 36.3 \\
R1-Distill-Llama-8B & 497 & 19.5 & 28.1 & 19.7 & 28.8 & 23.1 & 31.7 \\
\bottomrule
\end{tabular}
\caption{Normalized exact match (EM) and token F1 (\%) for CONFLICTS and NQ Open. Each model row uses its complete matched question set.}
\label{tab:standard_metrics}
\end{table*}

\subsection{RAMDocs Matching and Answer Instruction}
Table~\ref{tab:ramdocs_directional_sensitivity} applies directional all-gold matching on RAMDocs, which requires every normalized gold string to occur inside the prediction and keeps wrong-answer rejection. The point-estimate ordering between \mpdrel{} and \scscoring{} is unchanged for all five models, and their paired 95\% intervals all include zero.

Most RAMDocs questions have more than one gold answer, whereas the shared answer instruction asks for a short answer. We therefore also generate the Mistral-7B answers with an instruction that asks for every answer the documents support, separated by semicolons. Accuracy rises to 19.6 for \nf{}, 23.0 for \mpdrel{}, 21.2 when the selected documents keep their initial order, and 21.2 for \scscoring{}, so relevance scoring remains the most accurate recipe.

\begin{table}[t]
\centering\small
\setlength{\tabcolsep}{4pt}
\begin{tabular}{lccc}
\toprule
\textbf{Model} & \nf{} & \mpdrel{} & \scscoring{} \\
\midrule
Mistral-7B & 17.4 & 19.8 & 18.2 \\
Llama-3.1-70B-AWQ & 19.6 & 20.8 & 20.6 \\
Qwen3-4B-Instruct-2507 & 16.6 & 18.2 & 18.0 \\
Qwen3-8B & 13.0 & 12.6 & 12.8 \\
R1-Distill-Llama-8B & 12.0 & 12.8 & 11.8 \\
\bottomrule
\end{tabular}
\caption{RAMDocs accuracy (\%) under directional all-gold matching, which requires every normalized gold string inside the prediction and keeps wrong-answer rejection.}
\label{tab:ramdocs_directional_sensitivity}
\end{table}

\subsection{Controlled Distractor Metrics}
Table~\ref{tab:controlled_standard_metrics} reports all four controlled recipes under both distractor constructions, and Table~\ref{tab:controlled_effects} gives their paired effects. Every assessed recipe improves over \nf{} under every metric in both conditions. With other-question passages, the supplied evidence passages make up 46\% of the selected documents under \nf{} and between 74\% and 78\% under every assessed recipe and run. With one judgment, the quality instruction exceeds the relevance instruction under AM only with generated distractors, and exact match and token F1 show no such difference. Interactions subtract the other-question contrast from its generated-distractor counterpart, and the resamples share question indices across all eight cells.

\begin{table}[t]
\centering\small
\setlength{\tabcolsep}{4pt}
\begin{tabular}{lcccccc}
\toprule
& \multicolumn{3}{c}{\textbf{Generated}} & \multicolumn{3}{c}{\textbf{Other-question}} \\
\cmidrule(lr){2-4}\cmidrule(lr){5-7}
\textbf{Recipe} & \textbf{AM} & \textbf{EM} & \textbf{F1} & \textbf{AM} & \textbf{EM} & \textbf{F1} \\
\midrule
\nf{} & 41.6 & 26.4 & 35.7 & 71.4 & 59.0 & 67.5 \\
\rowcolor{methodblue}\mpdrel{} & 47.8 & 29.4 & 38.8 & 78.2 & 65.4 & 73.5 \\
\qualityonce{} & 51.4 & 29.2 & 39.8 & 77.2 & 64.8 & 73.1 \\
\scscoring{} & 51.0 & 30.8 & 40.4 & 76.8 & 65.2 & 73.1 \\
\bottomrule
\end{tabular}
\caption{Answer metrics (\%) for the controlled distractor comparisons on the same 500 questions, with generated distractors or passages of other questions.}
\label{tab:controlled_standard_metrics}
\end{table}

\begin{table*}[t]
\centering\small
\setlength{\tabcolsep}{4pt}
\begin{tabular}{llccc}
\toprule
\textbf{Contrast} & \textbf{Metric} & \textbf{Generated} & \textbf{Other-question} & \textbf{Interaction} \\
\midrule
\rowcolor{groupyellow}\multicolumn{5}{c}{\textit{Assessment recipes minus \nf{}}} \\
\mpdrel{} $-$ \nf{} & AM & +6.2 [+2.2, +10.4] & +6.8 [+4.0, +9.8] & $-$0.6 [$-$5.4, +4.2] \\
\qualityonce{} $-$ \nf{} & AM & +9.8 [+5.4, +14.0] & +5.8 [+2.6, +9.0] & +4.0 [$-$1.2, +9.0] \\
\scscoring{} $-$ \nf{} & AM & +9.4 [+5.2, +13.6] & +5.4 [+2.2, +8.6] & +4.0 [$-$1.2, +9.2] \\
\rowcolor{groupyellow}\multicolumn{5}{c}{\textit{Quality instruction minus relevance instruction, one judgment}} \\
\qualityonce{} $-$ \mpdrel{} & AM & +3.6 [+0.6, +6.6] & $-$1.0 [$-$3.0, +1.0] & +4.6 [+1.0, +8.2] \\
 & EM & $-$0.2 [$-$2.8, +2.6] & $-$0.6 [$-$2.4, +1.2] & +0.4 [$-$3.0, +3.6] \\
 & F1 & +1.0 [$-$1.4, +3.4] & $-$0.3 [$-$2.0, +1.4] & +1.3 [$-$1.8, +4.3] \\
\rowcolor{groupyellow}\multicolumn{5}{c}{\textit{Three quality draws minus the first draw}} \\
\scscoring{} $-$ \qualityonce{} & AM & $-$0.4 [$-$2.2, +1.4] & $-$0.4 [$-$1.8, +1.0] & 0.0 [$-$2.2, +2.2] \\
 & EM & +1.6 [+0.0, +3.2] & +0.4 [$-$0.8, +1.6] & +1.2 [$-$0.8, +3.2] \\
 & F1 & +0.6 [$-$0.8, +2.1] & 0.0 [$-$1.0, +0.9] & +0.6 [$-$1.1, +2.4] \\
\bottomrule
\end{tabular}
\caption{Controlled recipe differences with paired 95\% bootstrap intervals in points. Interaction is the generated-distractor contrast minus the other-question contrast.}
\label{tab:controlled_effects}
\end{table*}

\subsection{Paired Comparisons}
For a paired correctness comparison, let $d_i\in\{-1,0,1\}$ be the difference between two recipes on question $i$. We report $100\bar d$ and a 95\% percentile interval from 10{,}000 paired resamples of questions. Exact two-sided probabilities use the binomial test on helped versus hurt questions, which is the exact McNemar test without continuity correction. Tests with no discordant outcomes have $p=1$. Differences are computed before rounding. The percentile interval and the exact test rest on different reference distributions, so when few questions are discordant, an interval can exclude zero while the exact probability is at least .05.

Tables~\ref{tab:stats_primary_compact} and~\ref{tab:stats_branch_gaps} report the fifteen primary blocks. All probabilities are unadjusted, and the pooled contrasts of Appendix~\ref{app:pooled} summarize each benchmark across the five models. Tables~\ref{tab:stats_supplementary_compact} and~\ref{tab:stats_heldout} cover the controls, budgets, and held-out benchmarks, and Table~\ref{tab:stats_rankers} covers the single-pass rankers.

\begin{table*}[t]
\centering\small
\setlength{\tabcolsep}{5pt}
\begin{tabular}{lcccc}
\toprule
 & \multicolumn{2}{c}{\textbf{\mpdrel{} minus \nf{}}} & \multicolumn{2}{c}{\textbf{\scscoring{} minus \nf{}}} \\
\cmidrule(lr){2-3}\cmidrule(lr){4-5}
\textbf{Assessor and answerer} & $\Delta$ \textbf{[95\% interval]} & $p$ & $\Delta$ \textbf{[95\% interval]} & $p$ \\
\midrule
\rowcolor{groupyellow}\multicolumn{5}{c}{\textit{CONFLICTS}} \\
Mistral-7B & $-$2.1 [$-$7.2, +2.5] & .500 & $-$3.0 [$-$7.2, +1.3] & .265 \\
Llama-3.1-70B-AWQ & +6.8 [+3.0, +10.5] & $<.001$ & +5.5 [+1.3, +9.7] & .024 \\
Qwen3-4B & +5.5 [+0.8, +10.5] & .041 & +5.1 [+0.4, +9.7] & .050 \\
Qwen3-8B & +3.8 [$-$0.4, +8.4] & .136 & +1.3 [$-$3.8, +6.3] & .736 \\
R1-Distill-Llama-8B & +2.5 [$-$3.0, +8.0] & .451 & +5.1 [+0.0, +10.5] & .088 \\
\rowcolor{groupyellow}\multicolumn{5}{c}{\textit{RAMDocs}} \\
Mistral-7B & +2.4 [+0.2, +4.6] & .050 & +1.0 [$-$1.4, +3.4] & .522 \\
Llama-3.1-70B-AWQ & +1.2 [+0.0, +2.6] & .146 & +0.8 [$-$1.2, +2.8] & .557 \\
Qwen3-4B-Instruct-2507 & +1.6 [$-$0.2, +3.4] & .115 & +1.4 [$-$0.2, +3.0] & .143 \\
Qwen3-8B & $-$0.4 [$-$1.0, +0.0] & .500 & $-$0.2 [$-$1.0, +0.6] & 1.000 \\
R1-Distill-Llama-8B & +0.4 [$-$1.4, +2.4] & .839 & +0.2 [$-$1.6, +2.0] & 1.000 \\
\rowcolor{groupyellow}\multicolumn{5}{c}{\textit{NQ Open}} \\
Mistral-7B & +1.2 [$-$1.8, +4.2] & .504 & +3.0 [+0.0, +6.0] & .072 \\
Llama-3.1-70B-AWQ & +2.0 [$-$0.4, +4.4] & .154 & +2.2 [$-$0.4, +4.8] & .144 \\
Qwen3-4B-Instruct-2507 & +3.8 [+0.6, +6.8] & .025 & +4.6 [+1.6, +7.8] & .006 \\
Qwen3-8B & +2.0 [$-$1.0, +5.0] & .253 & +2.8 [$-$0.4, +6.0] & .120 \\
R1-Distill-Llama-8B & +4.4 [+1.4, +7.4] & .007 & +6.2 [+3.0, +9.5] & $<.001$ \\
\bottomrule
\end{tabular}
\caption{Primary effects against \nf{} in accuracy points, with paired bootstrap intervals and unadjusted exact paired probabilities.}
\label{tab:stats_primary_compact}
\label{tab:stats_conflicts}
\label{tab:stats_ramdocs}
\label{tab:stats_nq}
\end{table*}

\begin{table*}[t]
\centering\small
\setlength{\tabcolsep}{6pt}
\begin{tabular}{lcccc}
\toprule
\textbf{Assessor and answerer} & $\Delta$ & \textbf{Help/hurt} & \textbf{95\% interval} & $p$ \\
\midrule
\rowcolor{groupyellow}\multicolumn{5}{c}{\textit{CONFLICTS}} \\
Mistral-7B & +0.8 & 10/8 & [$-$2.5, +4.2] & .815 \\
Llama-3.1-70B-AWQ & +1.3 & 6/3 & [$-$1.3, +3.8] & .508 \\
Qwen3-4B & +0.4 & 6/5 & [$-$2.1, +3.0] & 1.000 \\
Qwen3-8B & +2.5 & 12/6 & [$-$0.8, +5.9] & .238 \\
R1-Distill-Llama-8B & $-$2.5 & 13/19 & [$-$7.2, +2.1] & .377 \\
\rowcolor{groupyellow}\multicolumn{5}{c}{\textit{RAMDocs}} \\
Mistral-7B & +1.4 & 19/12 & [$-$0.8, +3.6] & .281 \\
Llama-3.1-70B-AWQ & +0.4 & 8/6 & [$-$1.0, +1.8] & .791 \\
Qwen3-4B-Instruct-2507 & +0.2 & 4/3 & [$-$0.8, +1.2] & 1.000 \\
Qwen3-8B & $-$0.2 & 1/2 & [$-$0.8, +0.4] & 1.000 \\
R1-Distill-Llama-8B & +0.2 & 12/11 & [$-$1.6, +2.2] & 1.000 \\
\rowcolor{groupyellow}\multicolumn{5}{c}{\textit{NQ Open}} \\
Mistral-7B & $-$1.8 & 14/23 & [$-$4.2, +0.6] & .188 \\
Llama-3.1-70B-AWQ & $-$0.2 & 9/10 & [$-$2.0, +1.4] & 1.000 \\
Qwen3-4B-Instruct-2507 & $-$0.8 & 10/14 & [$-$2.8, +1.0] & .541 \\
Qwen3-8B & $-$0.8 & 18/22 & [$-$3.2, +1.6] & .636 \\
R1-Distill-Llama-8B & $-$1.8 & 26/35 & [$-$4.8, +1.2] & .306 \\
\bottomrule
\end{tabular}
\caption{Relevance scoring minus repeated quality scoring in the fifteen primary blocks, in accuracy points with paired bootstrap intervals and exact paired probabilities.}
\label{tab:stats_branch_gaps}
\end{table*}

\begin{table*}[t]
\centering\small
\setlength{\tabcolsep}{6pt}
\begin{tabular}{lcccc}
\toprule
\textbf{Comparison} & $\Delta$ & \textbf{Help/hurt} & \textbf{95\% interval} & $p$ \\
\midrule
\rowcolor{groupyellow}\multicolumn{5}{c}{\textit{Additional judgments and roles on CONFLICTS, Mistral-7B, $k=3$}} \\
\relonce{} $-$ \mpdrel{} & $-$1.7 & 4/8 & [$-$4.6, +1.3] & .388 \\
\screl{} $-$ \relonce{} & +0.4 & 6/5 & [$-$2.1, +3.4] & 1.000 \\
\qualityonce{} $-$ \mpdrel{} & +0.8 & 14/12 & [$-$3.4, +5.1] & .845 \\
\scscoring{} $-$ \qualityonce{} & $-$0.4 & 3/4 & [$-$2.5, +1.7] & 1.000 \\
\mpdthree{} $-$ \mpdrel{} & +2.1 & 12/7 & [$-$1.3, +5.9] & .359 \\
\scanswer{} $-$ \nf{} & 0.0 & 3/3 & [$-$2.1, +2.1] & 1.000 \\
\rowcolor{groupyellow}\multicolumn{5}{c}{\textit{Additional judgments and roles on RAMDocs, Mistral-7B, $k=3$}} \\
\relonce{} $-$ \mpdrel{} & $-$0.8 & 0/4 & [$-$1.6, $-$0.2] & .125 \\
\screl{} $-$ \relonce{} & 0.0 & 3/3 & [$-$1.0, +1.0] & 1.000 \\
\qualityonce{} $-$ \mpdrel{} & $-$1.4 & 8/15 & [$-$3.4, +0.4] & .210 \\
\scscoring{} $-$ \qualityonce{} & +0.6 & 7/4 & [$-$0.6, +2.0] & .549 \\
\mpdthree{} $-$ \mpdrel{} & $-$0.6 & 13/16 & [$-$2.8, +1.4] & .711 \\
\scanswer{} $-$ \nf{} & 0.0 & 7/7 & [$-$1.4, +1.4] & 1.000 \\
\rowcolor{groupyellow}\multicolumn{5}{c}{\textit{Additional judgments and roles on CONFLICTS, Qwen3-8B}} \\
\relonce{} $-$ \mpdrel{}, $k=3$ & $-$0.8 & 0/2 & [$-$2.1, 0.0] & .500 \\
\relonce{} $-$ \mpdrel{}, $k=5$ & +0.4 & 1/0 & [+0.0, +1.3] & 1.000 \\
\screl{} $-$ \relonce{}, $k=3$ & +0.4 & 2/1 & [$-$0.8, +2.1] & 1.000 \\
\screl{} $-$ \relonce{}, $k=5$ & 0.0 & 1/1 & [$-$1.3, +1.3] & 1.000 \\
\qualityonce{} $-$ \mpdrel{}, $k=3$ & $-$1.3 & 8/11 & [$-$5.1, +2.1] & .648 \\
\qualityonce{} $-$ \mpdrel{}, $k=5$ & $-$2.5 & 7/13 & [$-$6.3, +1.3] & .263 \\
\scscoring{} $-$ \qualityonce{}, $k=3$ & $-$0.4 & 0/1 & [$-$1.3, 0.0] & 1.000 \\
\scscoring{} $-$ \qualityonce{}, $k=5$ & $-$0.4 & 3/4 & [$-$2.5, +1.7] & 1.000 \\
\mpdthree{} $-$ \mpdrel{}, $k=3$ & $-$1.3 & 10/13 & [$-$5.1, +2.5] & .678 \\
\mpdthree{} $-$ \mpdrel{}, $k=5$ & $-$2.1 & 6/11 & [$-$5.5, +1.3] & .332 \\
\scanswer{} $-$ \nf{}, $k=3$ & $-$0.4 & 0/1 & [$-$1.3, 0.0] & 1.000 \\
\scanswer{} $-$ \nf{}, $k=5$ & 0.0 & 1/1 & [$-$1.3, +1.3] & 1.000 \\
\rowcolor{groupyellow}\multicolumn{5}{c}{\textit{Document budget, relevance scoring minus \nf{}}} \\
CONFLICTS, Mistral-7B, $k=3$ & +7.6 & 29/11 & [+2.5, +12.7] & .006 \\
CONFLICTS, Mistral-7B, $k=10$ & +1.3 & 12/9 & [$-$2.5, +5.1] & .664 \\
CONFLICTS, Llama-3.1-70B, $k=3$ & +11.4 & 34/7 & [+6.3, +16.5] & $<.001$ \\
CONFLICTS, Llama-3.1-70B, $k=10$ & +2.1 & 7/2 & [$-$0.4, +4.6] & .180 \\
CONFLICTS, Qwen3-8B, $k=3$ & +13.5 & 38/6 & [+8.4, +19.0] & $<.001$ \\
RAMDocs, Mistral-7B, $k=3$ & +2.8 & 23/9 & [+0.6, +5.0] & .020 \\
RAMDocs, Llama-3.1-70B, $k=3$ & +1.2 & 11/5 & [$-$0.4, +2.8] & .210 \\
\rowcolor{groupyellow}\multicolumn{5}{c}{\textit{Selection by \mpdrel{} in the initial order minus \nf{}, Mistral-7B}} \\
CONFLICTS, $k=3$ & +8.9 & 32/11 & [+3.4, +14.3] & .002 \\
CONFLICTS, $k=5$ & $-$0.4 & 17/18 & [$-$5.5, +4.6] & 1.000 \\
NQ Open & +3.6 & 33/15 & [+1.0, +6.4] & .013 \\
RAMDocs & +1.8 & 12/3 & [+0.4, +3.4] & .035 \\
RAMDocs, $N>5$ & +3.8 & 12/3 & [+0.8, +7.1] & .035 \\
\rowcolor{groupyellow}\multicolumn{5}{c}{\textit{Score order minus initial order after selection by \mpdrel{}, Mistral-7B}} \\
CONFLICTS, $k=3$ & $-$1.3 & 1/4 & [$-$3.0, +0.4] & .375 \\
CONFLICTS, $k=5$ & $-$1.7 & 3/7 & [$-$4.2, +0.8] & .344 \\
NQ Open & $-$2.4 & 6/18 & [$-$4.4, $-$0.6] & .023 \\
RAMDocs & +0.6 & 12/9 & [$-$1.2, +2.4] & .664 \\
RAMDocs, $N>5$ & 0.0 & 5/5 & [$-$2.5, +2.5] & 1.000 \\
\bottomrule
\end{tabular}
\caption{Paired effects for the additional judgments and roles, the document budgets, and the order controls, in accuracy points with bootstrap intervals and exact paired probabilities.}
\label{tab:stats_supplementary_compact}
\label{tab:stats_budget}
\end{table*}

\begin{table*}[t]
\centering\small
\setlength{\tabcolsep}{6pt}
\begin{tabular}{lcccc}
\toprule
\textbf{Comparison} & $\Delta$ & \textbf{Help/hurt} & \textbf{95\% interval} & $p$ \\
\midrule
\rowcolor{groupyellow}\multicolumn{5}{c}{\textit{Sampled recipes on NQ Open, Mistral-7B}} \\
\scscoring{} in the initial order $-$ \nf{} & +3.8 & 37/18 & [+1.0, +6.8] & .014 \\
\qualityonce{} in the initial order $-$ \nf{} & +3.8 & 37/18 & [+1.0, +6.6] & .014 \\
\scscoring{} in score order $-$ initial order & $-$0.8 & 13/17 & [$-$3.0, +1.2] & .585 \\
\qualityonce{} in score order $-$ initial order & +1.6 & 20/12 & [$-$0.6, +3.8] & .215 \\
\qualityonce{} $-$ \mpdrel{}, initial order & +0.2 & 17/16 & [$-$2.0, +2.4] & 1.000 \\
\scscoring{} $-$ \mpdrel{}, initial order & +0.2 & 14/13 & [$-$1.8, +2.2] & 1.000 \\
\qualityonce{} $-$ \mpdrel{}, score order & +4.2 & 31/10 & [+1.8, +6.6] & .001 \\
\scscoring{} $-$ \qualityonce{}, score order & $-$2.4 & 4/16 & [$-$4.2, $-$0.8] & .012 \\
\rowcolor{groupyellow}\multicolumn{5}{c}{\textit{Sampled recipes on NQ Open, Llama-3.1-70B}} \\
\scscoring{} $-$ \qualityonce{}, score order & +0.2 & 8/7 & [$-$1.4, +1.8] & 1.000 \\
\scscoring{} $-$ \qualityonce{}, initial order & +0.8 & 8/4 & [$-$0.6, +2.2] & .388 \\
\rowcolor{groupyellow}\multicolumn{5}{c}{\textit{MuSiQue}} \\
\mpdrel{} $-$ \nf{} & +12.4 & 84/22 & [+8.6, +16.4] & $<.001$ \\
\scscoring{} $-$ \nf{} & +14.0 & 94/24 & [+10.0, +18.0] & $<.001$ \\
\scscoring{} $-$ \mpdrel{} & +1.6 & 38/30 & [$-$1.6, +4.8] & .396 \\
\qualityonce{} $-$ \nf{} & +12.8 & 92/28 & [+8.8, +17.0] & $<.001$ \\
\scscoring{} $-$ \qualityonce{} & +1.2 & 29/23 & [$-$1.6, +4.0] & .488 \\
\rowcolor{groupyellow}\multicolumn{5}{c}{\textit{AmbigQA}} \\
\mpdrel{} $-$ \nf{} & +7.4 & 60/23 & [+4.0, +11.0] & $<.001$ \\
\scscoring{} $-$ \nf{} & +6.8 & 63/29 & [+3.2, +10.4] & $<.001$ \\
\scscoring{} $-$ \mpdrel{} & $-$0.6 & 19/22 & [$-$3.2, +1.8] & .755 \\
\qualityonce{} $-$ \nf{} & +7.6 & 63/25 & [+4.0, +11.4] & $<.001$ \\
\scscoring{} $-$ \qualityonce{} & $-$0.8 & 14/18 & [$-$3.0, +1.4] & .597 \\
\mpdrel{} $-$ padding-skip rule & +5.4 & 50/23 & [+2.2, +8.6] & .002 \\
\scscoring{} $-$ padding-skip rule & +4.8 & 53/29 & [+1.2, +8.4] & .011 \\
\bottomrule
\end{tabular}
\caption{Paired effects for the sampled recipes on NQ Open and the two held-out benchmarks, in accuracy points with bootstrap intervals and exact paired probabilities.}
\label{tab:stats_heldout}
\end{table*}

\begin{table*}[t]
\centering\small
\setlength{\tabcolsep}{5pt}
\begin{tabular}{llcccccc}
\toprule
& & \multicolumn{3}{c}{\textbf{Ranker minus \nf{}}} & \multicolumn{3}{c}{\textbf{Ranker minus \mpdrel{}}} \\
\cmidrule(lr){3-5}\cmidrule(lr){6-8}
\textbf{Benchmark} & \textbf{Ranker} & $\Delta$ & \textbf{95\% interval} & $p$ & $\Delta$ & \textbf{95\% interval} & $p$ \\
\midrule
CONFLICTS & MiniLM & +2.5 & [$-$2.1, +7.2] & .392 & +4.6 & [+0.0, +9.3] & .071 \\
 & BGE & +2.1 & [$-$2.5, +6.3] & .458 & +4.2 & [$-$0.4, +8.9] & .099 \\
 & RankGPT & +0.4 & [$-$4.2, +5.1] & 1.000 & +2.5 & [$-$2.1, +7.2] & .377 \\
RAMDocs & MiniLM & +0.2 & [$-$2.4, +2.8] & 1.000 & $-$2.2 & [$-$4.6, +0.2] & .090 \\
 & BGE & +2.2 & [$-$0.4, +4.8] & .135 & $-$0.2 & [$-$3.0, +2.6] & 1.000 \\
 & RankGPT & $-$1.2 & [$-$3.8, +1.4] & .441 & $-$3.6 & [$-$6.2, $-$1.0] & .010 \\
NQ Open & MiniLM & +2.6 & [$-$0.6, +5.8] & .124 & +1.4 & [$-$1.6, +4.4] & .443 \\
 & BGE & +3.6 & [+0.4, +6.8] & .033 & +2.4 & [$-$0.4, +5.4] & .141 \\
 & RankGPT & +3.2 & [+0.2, +6.4] & .056 & +2.0 & [$-$0.8, +5.0] & .220 \\
\bottomrule
\end{tabular}
\caption{Paired effects of the single-pass rankers with Mistral-7B, in accuracy points with bootstrap intervals and exact paired probabilities.}
\label{tab:stats_rankers}
\label{tab:stats_rerankers}
\end{table*}

\subsection{Pooled Contrasts}
\label{app:pooled}
Table~\ref{tab:pooled_contrasts} pools each benchmark over the five models. The pooled contrast averages the paired accuracy differences of the five models with equal weight. Each of 10{,}000 bootstrap draws samples, with replacement, as many questions as the union of the five question sets holds, and each model averages its differences over the drawn questions that it answered, so a question shared by several models moves all of their differences together. The fourth row sends the documents selected on NQ Open in their initial order, and the last row shuffles every RAMDocs pool before selection, with one order shared by the five models. Each judgment reads one document, so shuffling keeps the scores and changes only the documents \nf{} sends, the tie order, and the context order. Relative to the released order, shuffling changes the accuracy of \nf{} by $-$1.40 points [$-$2.64, $-$0.16] and the gain of relevance scoring by +0.20 [$-$0.76, +1.16].

\begin{table*}[t]
\centering\small
\setlength{\tabcolsep}{6pt}
\begin{tabular}{lcccc}
\toprule
\textbf{Benchmark} & \textbf{Questions} & \textbf{\mpdrel{} minus \nf{}} & \textbf{\scscoring{} minus \nf{}} & \textbf{\mpdrel{} minus \scscoring{}} \\
\midrule
CONFLICTS & 237 & +3.29 [+0.34, +6.41] & +2.78 [$-$0.34, +5.99] & +0.51 [$-$1.10, +2.11] \\
RAMDocs & 500 & +1.04 [+0.28, +1.84] & +0.64 [$-$0.20, +1.52] & +0.40 [$-$0.24, +1.04] \\
NQ Open & 618 & +2.69 [+1.06, +4.39] & +3.77 [+1.99, +5.59] & $-$1.08 [$-$2.12, $-$0.04] \\
NQ Open, initial order & 618 & +2.81 [+1.15, +4.55] & +3.61 [+1.94, +5.34] & $-$0.81 [$-$1.69, +0.07] \\
RAMDocs, shuffled & 500 & +1.24 [+0.40, +2.12] & +1.20 [+0.16, +2.24] & +0.04 [$-$0.60, +0.72] \\
\bottomrule
\end{tabular}
\caption{Differences pooled over the five models in accuracy points, with 95\% question-cluster bootstrap intervals. Questions counts the union of the question sets of the five models, and the last two rows change the document order.}
\label{tab:pooled_contrasts}
\end{table*}

\end{document}